\documentclass[journal,12pt,onecolumn]{IEEEtran}

\IEEEoverridecommandlockouts
\usepackage{cite}
\usepackage{amsmath,amssymb,amsfonts}
\usepackage{algorithmic}
\usepackage{graphicx}
\usepackage{textcomp}
\usepackage{xcolor}
\usepackage{algorithm2e}
\usepackage{array}
\newcolumntype{P}[1]{>{\centering\arraybackslash}p{#1}}
\usepackage{amsmath} 
\usepackage{amsthm} 

\usepackage[numbers,sectionbib]{natbib}
\def\BibTeX{{\rm B\kern-.05em{\sc i\kern-.025em b}\kern-.08em
    T\kern-.1667em\lower.7ex\hbox{E}\kern-.125emX}}
    
\newtheorem{thm} {Theorem} 
\newcommand {\BT} {\begin{thm}}
\newcommand {\ET} {\end{thm}}  
\usepackage{float}

\begin{document}

\title{Locality Sensitive Hashing-based Sequence Alignment Using Deep Bidirectional LSTM Models}
%
%
\author{Neda~Tavakoli
\thanks{N. Tavakoli is with the Department
of Computer Science, Georgia Institute of Technology, Atlanta,
GA, 30332 USA e-mail: neda.tavakoli@gatech.edu}

\thanks{}}

\maketitle
%
\begin{abstract}
Bidirectional Long Short-Term Memory (LSTM) is a special kind of Recurrent Neural Network (RNN) architecture which is designed to model sequences and their long-range dependencies more precisely than RNNs. This paper proposes to use deep bidirectional LSTM for sequence modeling as an approach to perform locality-sensitive hashing (LSH)-based sequence alignment. In particular, we use the deep bidirectional LSTM to learn features of LSH. The obtained LSH is then can be utilized to perform sequence alignment. We demonstrate the feasibility of the modeling sequences using the proposed LSTM-based model by aligning the short read queries over the reference genome. We use the human reference genome as our training dataset, in addition to a set of short reads generated using Illumina sequencing technology. The ultimate goal is to align query sequences into a reference genome. We first decompose the reference genome into multiple sequences. These sequences are then fed into the bidirectional LSTM model and then mapped into fixed-length vectors. These vectors are what we call the trained LSH, which can then be used for sequence alignment. The case study shows that using the introduced LSTM-based model, we achieve higher accuracy with the number of epochs.  

{\bf Keywords}: Bidirectional Long short-term memory (LSTM), sequence alignment, locality-sensitive hashing (LSH), recurrent neural network (RNN). 
\end{abstract}


\section{Introduction}
\label{Introduction}

The emerging technology in High Throughput Sequencing (HTS) such as Illumina~\cite{bennett2004solexa}, SOLiD (Applied Biosystems)~\cite{ondov2008efficient}, and the 454 pyrosequencing has rapidly increased the needs for designing algorithms that efficiently map huge amounts of reads to a reference genome. NGS (Next Generation Sequencing) generates a huge amount of DNA sequencing (reads) in short time at low cost. However, analyzing these substantial amount of reads in a linear time is a computationally challenging problem. In particular, sequence alignment of a long query sequence over a large reference genome is computationally intensive. To date, several algorithmic methods have been proposed to improve the performance of the existing sequence alignment tools for long and short reads. For example, parallel algorithms have been designed to execute alignment tools on high performance computing architectures~\cite{liu2012soap3}. As an example, SOAP3~\cite{liu2012soap3} is the first short read alignment tool that leverages the high performance computing architecture in Graphic Processing Unit (GPU) to achieve a drastic improvement at run-time. Other research studies have also leveraged GPUs and Field Programmable Gate Arrays (FPGAs) to accelerate the run-time of the alignment tools~\cite{mckenna2010genome}.

In this work, we leverage the Recurrent Neural Network (RNN), more specifically the deep bidirectional Long Short-Term Memory (LSTM) model, and the GPU platforms to accelerate sequence alignment tools. 
%

RNNs are extremely powerful learning models that have demonstrated excellent success on addressing hard problems such as objection recognition
and speech recognition. Furthermore, RRNs can be used in many applications of Natural Language Processing (NLP) such as language modeling~,time series forecasting and clustering~\cite{siami2018comparison,BigData2019BiLSTM, Abrietal2019}, paraphrase detection ~\cite{socher2011dynamic}, genome modeling~\cite{li2019identifying, gupta2017dilated},
word embedding extraction~\cite{mikolov2013distributed},
and managing systems~\cite{tavakoli2019client,tavakoli2018software, tavakoli2016log}. 

These applications and the datasets are very similar to genome data in which the input data shape is in the form of strings. 
RNNs are powerful tools because they are able to perform parallel computation for a modest number of steps. On the other hand, RNNs are not capable of learning long-term dependencies. As a result, in this work we adapt a special type of RNNs,  called deep bidirectional Long Short Term Memory (LSTM), to address the problem of modeling long-reads. In~\cite{gupta2017dilated},authors showed that dilated convolutions can be used to capture long-range relationships in DNA.

The ultimate goal is to align query sequences into a reference genome. To this end, we first map the entire reference genome into multiple fixed-length vectors, which is closely related to the study performed by Kalchbrenner and Blunsom~\cite{kalchbrenner2013recurrent}. One of the main differences between the approach presented in this paper and the work reported in~\cite{kalchbrenner2013recurrent} is that they map the entire input sequence into a one vector. In this work, however, we first map the entire reference genome into multiple fixed-length vectors. 
These obtained vectors are considered as the trained model that learn the conventional Locality-Sensitive Hashing (LSH). Then, the  query sequences will be fed into the model to obtain a single vector representation of the sequences.
Finally, we compare the vector generated per query sequence with those of the reference sequence to find the best candidate vector, which is defined later in the process. 
The key contributions of this paper are:
\begin{enumerate}
    \item Introduce a novel learning algorithm using the bidirectional LSTM model to generate fixed-length vector representations of the reference sequence;
    \item Leverage the LSTM model to learn the Locality Sensitive Hashing (LSH), and perform sequence alignment;
    \item Conduct a case study on real DNA sequencing dataset, and compute the perplexity,  accuracy and sensitivity of the LSTM-based sequence alignment algorithm.
\end{enumerate}
 Section~\ref{Problem-definition} presents a formal definition of the problem. Section~\ref{background} provides a detailed background needed for this work such as RNN, LSTM and bidirectional LSTM, Locality Sensitive Hashing (LSH), and sequence alignment. Section~\ref{Training-Algorithm} describes the proposed LSTM-based training algorithm as well as leveraging the obtained trained model to perform LSH-based sequence alignment. Section~\ref{Experimental-Results} reports the results of case study in which the proposed modeling is assessed. We conclude the paper and discuss the future work in Section~\ref{Conclusion}.

\section{Problem Definition}\label{Problem-definition}
Let the sequence $G$ represents the genome of an organism, which is called the reference genome, with a length of $|G|$.  Suppose that $D =\{ S_1, S_2, S_3, \dots, S_n\}$ represents a set of $n$ sequences, not necessarily of the same length, with a total length of $N$ i.e., $|S_1| + |S_2| +|S_3|+\dots +|S_n| = N$.
The goal is to design an algorithm such that for all the sequences in  D, called the {\it query sequences}, report all substrings that align with at least one substring of $G$ with the highest alignment score. Specifically, the algorithm outputs a tuple $(q, r, t)$ per sequence in $D$ if the $S_q[r\dots r+l-1]$ and $G[t \dots t+l-1]$ are aligned together with the highest score ($l$ is the length of the alignment). This problem is called sequence alignment which will be disused in more details in Section~\ref{Sequence-alignment}. In this paper, the aforementioned problem is modeled using LSTM to learn a locality sensitive hashing used to align sequences.

\section{background}\label{background}
\label{back}
\subsection{Recurrent Neural Networks}\label{RNN}
In theory, RNNs can make use of sequential information for arbitrary long sequences. In practice, however, these sequential information are limited to look back only a few steps (i.e., the limited memory length). To have a formal definition of RNN, lets consider $x= (x_1, x_2, x_3, ....,x_T)$ to represent a sequence of length $T$, an RNN model updates its recurrent hidden state $h_t$ using the following formula:
\begin{equation} \label{eq1}
h_t = \sigma( W_x x_t + W_h h_{t-1} +b_t)
\end{equation}
where $\sigma$ is a nonlinear function such as logistic sigmoid function, a hyperbolic tangent function, or rectified linear unit (ReLU), $W_x$ and $W_h$ are weight matrices, and $b_t$ is a constant bias value.
RNNs can map an input to many outputs, many inputs to many outputs, and many inputs to one output. In this work, we consider RNNs that produce an output $y = (y_1, y_2, ...., y_T)$ of a probability distribution over the next element of the sequence while taking into consideration the current input and also what it has been learned from the inputs the model received previously. The sequence probability can be decomposed as following:
\begin{equation}\label{eq2}
\begin{split}
        p(x_1,\dots,x_T)& = p(x_1)  p(x_2 |x_1) p(x_3|x_1, x_2)...\\
        &~~~~p(x_T | x_1, .....,x_{T-1})
\end{split}
\end{equation}
where the last element is a special end-of-sequence value. Each conditional probability distribution is modeled as follows:
\begin{equation}\label{eq3}
p(x_t | x_1, \dots x_{t-1}) = \sigma (h_t)
\end{equation}
 where $h_t$ is calculated using Equation~\ref{eq1}. 
It is hard to train RNNs and capture long-term dependencies because the gradients tend to either vanish (most of the time) or explode (rarely, but with severe effects). 

There are two major problems with the standard RNNs: 1) ``{\it exploding gradients},'' and 2) ``{\it vanishing gradients}'', which leads to make it hard to train and to capture long-term dependencies. To explain these issues in further details, lets first explain the semantic of the gradient. The gradient measures how much the output of a function changes with respect to the changes occurred to its inputs (i.e., a partial derivative with respect to its inputs). Likewise, the {\it exploding gradients} problem refers to the case when the algorithm assigns high importance to the weight matrix without any reasons. This problem can be solved by squashing or truncating the gradients~\cite{hochreiter1997long}.
The other challenging issue with the standard RNNs, which is emerged as a major obstacle, is called vanishing gradients which occurs when the values of gradients are too small and thus the RNN model stops learning. Such problem can be solved using an LSTM model which will be discussed in Section ~\ref{lstm}.
\subsection{Long Short-Term Memory (LSTM) Models}\label{lstm}
The LSTM-based learning networks are an extension for RNNs. These models are capable of addressing the vanishing gradient problem in a very clean manner (i.e., RNN's difficulties in learning long-term dependencies). LSTM networks extend the RNNs memory and enable them learn long-term dependencies. They can remember information over a long period of time and can read, write, and delete information from theirs memories. The LSTM memory is called a ``{\it gated}'' cell, in which a gate refers to its ability to make the decision of preserving or ignoring the memory.

Intuitively, if an LSTM model captures important features from an input sequence, it easily maintains this information over a long period of time and distance, otherwise it will remove it from further analysis. The decision of keeping or deleting the information is mainly made based on the importance level assigned to the information through weights (discussed in more details later). Indeed, an LSTM model learns what information worth to maintain or ignore.  

An LSTM model consists of three gates: forget, input, and output gates. The forget gate controls whether or not the existing information will be remained in the cell, the input gate determines the extent to which the new information will be added into the cell, and the output gate makes decision of whether or not the existing value in the cell will be used to compute the output of the LSTM. Lets explain all these gates in further detail:
\begin{enumerate}
    \item {\it Forget Gate}: This gate is typically a sigmoid function which is used to decide what information need to be thrown away from the LSTM memory. The decision is made based on the value of $h_{t-1}$ and $x_t$. It produces $f_t$ as follows:
    \begin{equation}\label{eq4}
    f_t = \sigma (W_{f_h}[h_{t-1}], W_{f_x}[x_t], b_f)
    \end{equation}
    $f_t$ is a number between 0 and 1 where 0 indicates to completely get rid of the learned value and 1 implies to completely keep the value. $b_f$ is a constant value, which is called the {\it bias} value.
    \item {\it Input Gate}: This gate is used to make the decision of whether or not the new information is going to be stored in the LSTM memory. This gate itself consists of two layers: 1) a {\it sigmoid} layer, and 2) a ``$\tanh$'' layer. The sigmoid layer is responsible to decide which values need to be updated; whereas, the $\tanh$ layer is used to creates a vector of new candidate values that can be added to the LSTM memory. The following equations represent the output of these two layers:
    \begin{align}
    i_t     &= \sigma (W_{i_h}[h_{t-1}], W_{i_x}[x_t], b_i) \label{eq5}\\
    c_t^{\texttt{\char`\~}}   &= \tanh(W_{c_h}[h_{t-1}], W_{c_x}[x_t], b_c)\label{eq6}
    \end{align}
    Where $i_t$ represents the decision of which values need to be updated, and $c_t^{\texttt{\char`\~}}$ shows a vector of new candidate values that can be added to the LSTM memory. These two layers are combined to create an update for the LSTM memory. To do so, first the current value needs to be forgotten using the forget gate layer described earlier. This can be done by multiplying the old value (i.e., $c_{t-1}$) by $f_t$ followed by by adding the new candidate value $i_t * c_t^{\texttt{\char`\~}} $. The exact formula is represented through the following equation: 
    \begin{align}
    c_t = f_t * c_{t-1} + i_t * c_t^{\texttt{\char`\~}} \label{eq7}
    \end{align}

    where $f_t$ is a number between $0$ and $1$ obtained from the forget gate. Likewise, $0$ implies to get rid of the value and $1$ indicates to completely keep the value that represents the extent to which the old information (i.e., $c_{t-1}$) needs to be remained in the LSTM memory.

    \item {\it Output Gate}: This gate is used to make a decision based on the value of the output. To do so, first a sigmoid layer is executed to make a decision of what part of the LSTM memory is considered for the output. Then a $\tanh$ is applied to push the value between $-1$ and $1$. the result is then multiplied by the the output of sigmoid layer. Hence, only parts that have been decided will be reported to the output. The following equations show the exact math of the process: 
    \begin{align}
    o_t  &= \sigma (W_{o_h}[h_{t-1}], W_{o_x}[x_t], b_o) \label{eq8}\\ 
    h_t  &= o_t * \tanh (c_t) \label{eq9}
    \end{align}
    where $o_t$ represents the output value, and $h_t$ indicates the output value between $-1$ and $1$.
\end{enumerate}
  
\subsection{Deep Bidirectional LSTMs} 
\begin{figure}[!t]
\centering
 \includegraphics[width={0.8\textwidth}]{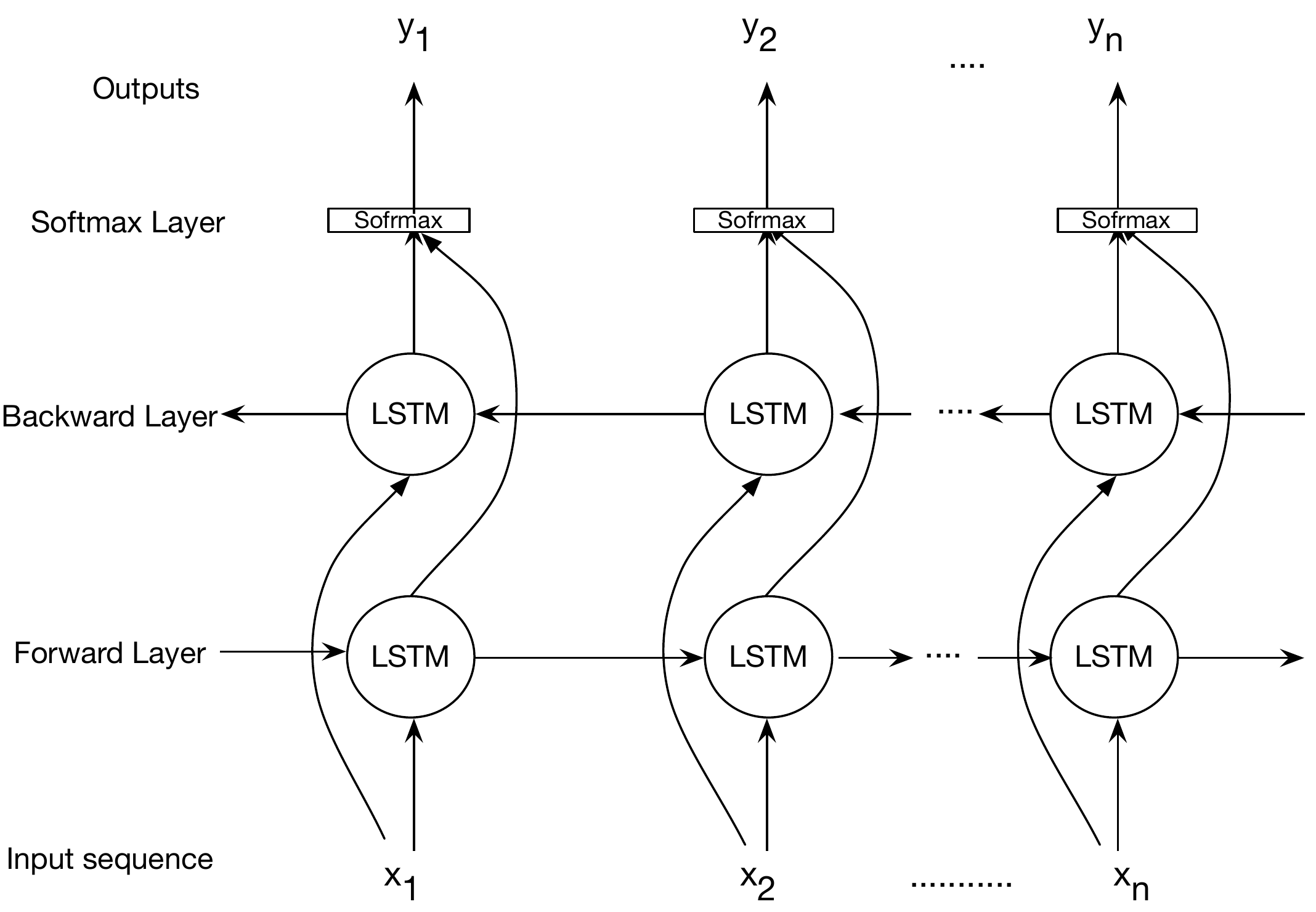} 
\caption{Bidirectional LSTM architecture.}
\label{fig2}
\end{figure}

Deep bidirectional LSTMs~\cite{schuster1997bidirectional} are an extended version of basic LSTMs where the trained model is obtained by applying LSTM twice. Once, the input sequences are fed as-is into the LSTM model (forward layer), a reversed version of the input sequences (i.e., Watson-Crick complement~\cite{gao2004expanded}) will be also fed to the LSTM model (backward layer). Using the bidirectional LSTMs can improve the performance of the model~\cite{baldi1999exploiting}. This paper uses bidirectional LSTMs to model genome data. Figure \ref{fig2} illustrates an architecture for bidirectional LSTM model employed in this paper.
\subsection{Locality-Sensitive Hashing (LSH)}\label{LSH}
Locality-Sensitive Hashing (LSH) is a family of functions, $F$, used to hash data items into ``buckets'' such that similar items are more likely mapped to the same buckets~\cite{indyk1998approximate}. If data items are far from each other they are likely mapped onto different buckets. Any pair that is mapped to the same bucket is called a candidate pair. 
%
The major difference between LSH and traditional hashing methods is about the ability of LSH to maximize the probability of collision for similar items.

To formally define locality-sensitive hashing, lets assume that the locality-sensitive function takes two data items and makes a decision of whether these items should be a candidate pair. The function $f\in F$ will hash each item, and the decision will be made on whether or not the results are equal. We use the notation $f(x) = f(y)$ to indicate that if $f(x, y) = yes$ and put the data items $x$ and $y$ into the same bucket, and vice versa. 

Lets $d_1 < d_2$ be two distances according to some distance metric $d$. A family of functions $F$ is defined to be $(d_1 , d_2, p_1, p_2)$-sensitive if for every function $f\in F$:  
\begin{enumerate}
    \item If $d(x, y) \leq d_1$, then the probability that $f(x) = f(y)$ is at least $p_1$.
    \item If $d(x, y) \ge d_2$, then the probability that $f(x) = f(y)$ is at most $p_2$.
\end{enumerate}
If the distance between the data items is strictly between $d_1$ and $d_2$, we can make $d_1$ and $d_2$ as close as we wish.



\subsection{Sequence Alignment}\label{Sequence-alignment}
Sequence alignment is an approach to represent the relationship between two sequences. Particularly, it arranges the sequences to identify regions of similarity among them. Briefly, to align two sequences together, a place of each symbol will be adjusted as follows:
\begin{itemize}
    \item Inserting some number of blanks in front of one sequence.
    \item Inserting some number of "-" symbol (a symbol which is not in the alphabet, and is also called gap) at some point in the sequence.
\end{itemize}
There exists two types of sequence alignments: {\it global} alignment, and {\it local} alignment. In global alignment, the goal is to align the entire sequence (i.e., end-to-end alignment). However, the local sequence alignment aims to align a substring of a query sequence to a substring of a reference sequence. 



\textbf{Formal Definition.}
Assume $x = (x_1, x_2, x_3, ....,x_T)$ represents a sequence of length $T$ of some symbols over a finite set of alphabet $\Sigma$, a very common example of $\Sigma$ includes the DNA alphabet $\{A, C, G, T\}$ or the $20$-letter alphabet of symbols used to represent the amino acids found in proteins. 
A {\it global alignment} A between two sequences $x = (x_1, x_2, x_3, ....,x_T)$ and $y = (y_1, y_2, y_3, ....,y_k)$ over the same alphabet $\Sigma$, consists of a pair of sequences $(x\prime, y\prime)$ of the same length, where $x\prime$ and $y\prime$ are the results of inserting zero or more special gap charterers (i.e., characters not found in $\Sigma$), in either $x$ or $y$. 
On the other hand, the {\it local alignment} refers to the global alignment of consecutive substrings of $x$ with those of $y$. 

The choice of local or global alignment is application dependent. For example, for short sequences, global alignment is appropriate. Conversely, for long sequences a collection of local alignments might be a better choice. 





%
\subsection{Alignment Score}\label{Alignment-score}
To measure the quality of an alignment for a given query sequence, the alignment score is defined as the summation of the symbol-wise score symbols, and the gap penalty. Symbol-wise score can be defined as a two-dimensional matrix, where rows and columns are characters, indicating score associated for aligning characters. The gap penalty can be computed based on the position of the gap. More specifically, if the gap appears in the leftmost of a block of gaps, we assign more penalty (here $-10$) in comparison to the other gaps. In general, computing alignment score can be done using a standard dynamic programming approach. A recurrence formula is needed to measure the score of the alignment, then a standard dynamic programming is used to compute the alignment score~\cite{smith1981identification}. An optimal alignment is defined as an alignment with the highest score.



\section{LSH-based Sequence Alignment Using LSTM}\label{Training-Algorithm}
\begin{figure}[!t]
\centering
  \includegraphics[width=1\textwidth]{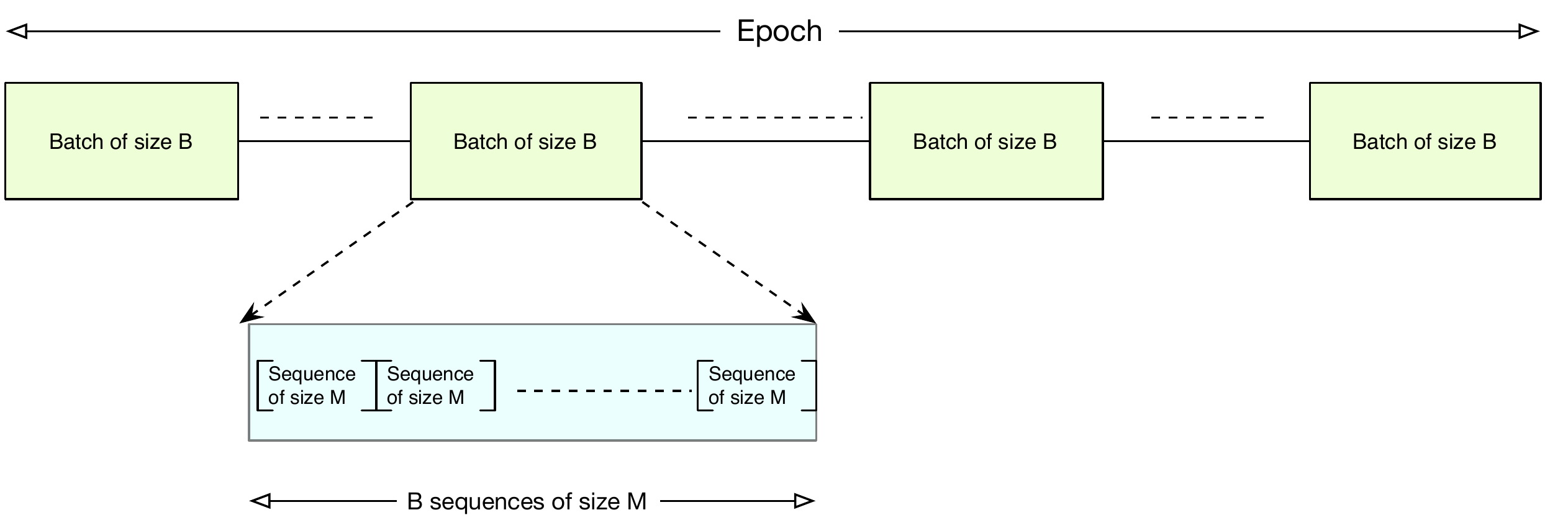}
\caption{Batches and Epoch inside the reference genome.}
\label{fig1}
\end{figure}

In this section, we explain our LSH-based sequence alignment using the LSTM model. To model sequences using the LSTM, the following training algorithm is used which learns the locality-sensitive hashing needed to perform sequence alignment algorithm.
\begin{itemize}
    \item \textbf{\texttt{Step 1}} {\it (Constructing Words).} Each word is made of $w$ characters, so the reference genome that includes a set of characters is split into unique words of size $w$ characters (where $w$ represents the word size). Hence, for the reference genome of size $|G|$, the number of unique words is at most $\frac{|G|}{w}$, and for the alphabet size of $\Sigma$, the max number of unique words is bounded by $|\Sigma|^w$.
    \item \textbf{\texttt{Step 2}} {\it (Constructing Dictionary).} Dictionary is defined as a collection of unique words. To construct a dictionary, each unique word, generated at the previous step, is added to the dictionary. The dictionary size is bounded by $|\Sigma|^w$.
    \item \textbf{\texttt{Step 3}} {\it (Constructing Batch).} Fig~\ref{fig1} illustrates the concept of batch, as the figure shows each batch is a collection of fixed-length words (words of length $w$). Indeed, the reference genome is split into multiple batches where each batch consists of multiple words. The rational behind using batch is that the memory of LSTM is limited, bounded by $M$ words, so every $M$ words are considered as a sequence which can be fed into the LSTM model. In a single batch, let assume there are $b$ sequences (of size $M$ words) therefore, $b$ parallel LSTM can run simultaneously on a single batch, one LSTM per sequence. The batch size, $B$, equals to $B = b \times M\times w$ words.
    The max number of batches inside a reference genome equals to $ \frac{|G|}{B}$. 
    As a result of feeding sequences to the LSTM model, hidden vectors are generated as mentioned earlier in Equation~\ref{eq9}, and finally the weight matrices of the LSTM model will be updated. This process will be repeated per batch. 
    \item \textbf{\texttt{Step 4}} {\it (Constructing Epoch (Iteration)).} Epoch is a collection of all batches. Each epoch runs over the whole reference genome. Hence, epoch size equals to $|G|$. Indeed, epoch consists of $ \frac{|G|}{B}$ batches. LSTM network runs batch by batch sequentially. The final updated weights are what is called the training model.
\end{itemize}
%
%
The results of training the reference genome using the above algorithm include multiple vectors (i.e., one vector per batch), each of dictionary size (i.e., at most $|\Sigma|^w$), where each vector generated sequentially. 
At the end, the updated weight matrix will be generated which is called the trained model. Finally, this trained model is used to perform sequence alignment. In particular, a query sequence of size $M$ words is fed into the LSTM model, and as a result a fixed-length vector representation of that sequence (i.e., hidden vector) will be generated. This vector then needs to compare with all the vectors generated from the reference genome. Section~\ref{LSTM_model for sequence alignment} will explain the algorithm in more details. 
%
 \subsection{LSTM Models for Sequence Alignment}\label{LSTM_model for sequence alignment}
The general algorithm of the LSTM-based sequence alignment is presented in Algorithm~\ref{alg:1}. 

\begin{algorithm}[H]
\label{alg:1}
\SetAlgoLined
\KwResult{The aligned sequences.}
  1) Decompose\;
  2) Feed Sequences to the LSTM Model (Phase 1)\;
  3) Decompose Query Sequences\;
  4) Feed Sequences to the LSTM Model (Phase 2)\;
  5) Compare Vectors\;
  6) Find the Best Candidate Vector\;
 \caption{LSTM-based sequence alignment.}
\end{algorithm}

Lets discuss each step of the algorithm in more details. Sequence alignment of a given query sequence over a reference genome using the LSTM model has the following steps:
\begin{enumerate}
    \item {\it Decomposing the Reference Genome.} The entire reference genome is divided into unique sequences of size $M$. There exists at most $|G| - M  +1 $ of such unique sequences inside the genome (extremely huge number of sequences).
    \item {\it Feeding Sequences to the LSTM Model (Phase 1).} Sequences obtained from the reference genome are fed into the trained LSTM model (one by one). A hidden vector is generated per sequence which indicates the fixed-length vector representation of that sequence.
    \item {\it Decomposing Query Sequences.} For all the given query sequences (without loss of generality, suppose they are of length $l$ s.t. $M \le l$), split them into words of size $M$.
    \item {\it Feeding Sequences to the LSTM Model (Phase 2).} Sequences obtained from any query sequence are fed into the trained LSTM model (one by one). A hidden vector is generated for each sequence which indicates the vector representation of that sequence.
    \item {\it Comparison Vectors.} Vectors generated for any query sequence need to be compared with those of generated for the reference genome. This step itself consists of several substeps as following (note that theses substeps need to be done for any query sequence):
        \begin{enumerate}
            \item Concatenate all the vectors generated for the query sequence and build one vector of size $\frac{l}{M} \times D$ ($l$ is the query length, each query consists of sequences of size $M$ s.t. $M\le l$ and $D$ is the dictionary size).
            \item Comparing vector representation of the query sequence to those of the reference genome leads to extremely huge search space which is quite expensive. Hence, a method is needed to reduce the search space. To do so, every collection of $\frac{l}{M} \times D$ sequences of size $M$ located in the reference genome are concatenated together to build a set of vectors of size $\frac{l}{M} \times D$.
            \item Vector of size $\frac{l}{M} \times D$ generated for the query sequence, needs be compared with those of the reference genome to find the best candidate vector to be aligned to (the best candidate vector is discussed in the next step).
            %
        \end{enumerate}
    %
    \item {\it Finding the Best Candidate Vector.} Through the concatenated vectors step, discussed in the previous step, we  reduce the search space an order of magnitude smaller than the previous. A measurement is needed to compare the vector representation of the sequence to to find the best candidate score. Thus, we compute the alignment score of each vector (as it was discussed in ~\ref{Alignment-score}), and at the end the vector with the maximum alignment score will be chosen as the best candidate vector.
\end{enumerate}
%

 

\section{Case Study}\label{Experimental-Results}
The feasibility of the proposed LSTM-based model is demonstrated through a case study in which sequences of one  chromosome are modeled. We developed several Python scripts to implement and assess the modeling aspect of the proposed LSTM-based sequence alignment algorithm. We evaluated the algorithm using human genome as the reference sequence and a set of short reads generated using Illumina sequencing technology
as the query sequences. More specifically, we used Genome Reference Consortium Human Build 38 patch release 13 (GRCh38.p13) as the reference genome~\cite{HumanG:2018},
which has a total sequence length of $3,099,734,149$. GRCh38.p13 consists of $24$ chromosomes with different lengths, for example its chromosome number $7$ has total length of $159,345,973$ base pairs (bp); whereas, its chromosome number $6$ has the total length of $170,805,979$ bp. 

For the query sequences, Illumina (HiSeq X Ten) short reads of length $151$ were used. 
Table~\ref{table1} shows the details for the query sequences, where ``Run'' represents the project ID. There exists two short reads (query sequences of length $151$) per spot (will be discussed later). Hence, for a total number of $401,747,042$ spots, the number of reads for this run is $2\times 401,747,042$.\\

\begin{table}[h!]
\centering
\caption{Short reads generated by Illumina.}
\label{table1}
\begin{tabular}{ |P{1.8cm}||P{1.8cm}|P{1cm}|P{1cm}| P{1cm}| }
 \hline
 \multicolumn{5}{|c|}{Query Sequences (Short Reads)} \\
 \hline
Run & Num. of Spots & Num. of Bases & Size & Read size\\
 \hline
SRR7205176  &	401,747,042    & 121.3G	 &   21Gb &   151 bp\\
 \hline
\end{tabular}
\end{table}
Short read query sequences are based pair reads which means that it has two reads per spot: 1) a forward read, and 2) a reverse biological read. For example, the read number $1$SRR$7205176.1$ consists of two parts (i.e., SRR$7205176.1.1 1 $ and SRR$7205176.1.2$). Table \ref{table2} lists an example of such reads.\\ 

\begin{table}[h!]
\centering
\caption{A query sequence of base pairs (151 characters).}
\label{table2}
\begin{tabular}{ |p{8.5cm}| }
 \hline
 \multicolumn{1}{|c|}{SRR$7205176.1$} \\
 \hline
\textbf{SRR7205176.1.1} (Biological, Forward)
TAACCCTAACCCTAACCCTAACCCTAACCCTAACCCTAACCCTAACCCTAACCTTAACCC
TAACCCTAACCCTAACCCTAACCCTAACCCTAACCCTAACCCTAACCCAACAACCAACCC
TAACCCTACCCCTACCCATCCACCTACCCCT\\
 \hline
\textbf{SRR7205176.1.2} (Biological, Reverse)
TTAGGGTTAGGGTTAGGGTTAGGGTTAGGGATAGGGTTAGGGTTAGGGTTAGGGTGGGGG
TTGGTTGTAGTGTTAGGGATAGGTGTCGGGTTAGGGTTAGGGTTTAAGTTGAGTAGCGGG
GTAGTTGTAGTATTTGTGATATCGCGTACTT\\
 \hline
\end{tabular}\\
\end{table}

All tests were run on the Swarm
cluster located at 
Each computer node in the cluster has dual Intel Xeon CPU E5-2680 v4 (2.40GHz) processors equipped with a total of 28 cores and 256GB main memory. The cluster is set up using 64-bit Red Hat Linux kernel version 2.6.32. Also, it is equipped with $4$ GPUs with Driver Version $418.40.04 $ and CUDA Version $10.1$. 

We trained the LSTM model by feeding the reference genome sequences (as we discussed earlier in Section~\ref{Training-Algorithm}), and obtained their hidden vectors representation as the result. Table~\ref{table3} lists the parameters and their corresponding values that we used to configure and train the LSTM-model.\\

\begin{table}[h!]
\centering
\caption{The configuration of the LSTM model.}
\label{table3}
\begin{tabular}{ |P{1.5cm}| P{1.5cm}| P{1.5cm}| P{2cm}| }
 \hline
 \multicolumn{4}{|c|}{Configuration of LSTM model} \\
 \hline
 \hline
Number of Layers & Hidden size & Word size & Words per Sequence\\
 \hline
1  & 1000  & 10  & 100 \\
\hline
\end{tabular}
\end{table}
Table ~\ref{table4} shows the results after training the model by feeding sequences of one chromosome of the reference genome. We ran the experiments for $100$ epochs and evaluated perplexity per epoch where perplexity is a measurement to show how well the LSTM model can predict the next word. Perplexity refers to the log-averaged inverse probability on unseen (i.e., test) data. A low perplexity indicates that the model is good at predicting the sample and its probability distribution. As Table~\ref{table4} indicates the perplexity values are decreasing with the increase of the number of epochs. Our results show that the perplexity values are reduced from $298.82$ to $20.10$ 
when the number of epochs has increased from $1$ to $100$ indicating the improvement in prediction for the given genome sequence.\\ 

%

\begin{table}[h!]
\centering
\caption{Training results of the LSTM model}
\label{table4}
\begin{tabular}{|P{1.3cm}| P{0.8cm}| P{0.8cm}| P{0.8cm}| P{0.4cm}| P{0.8cm}| P{0.8cm}|}
 \hline
 \multicolumn{7}{|c|}{Results of training the LSTM model} \\
 \hline
 \hline
Epoch number &  1&   2   &  3  & .... & 99 & 100 \\
 \hline
Perplexity & 298.82 & 270.77 & 259.42&.... & 23.26& 20.10\\
\hline
\end{tabular}\\
\end{table}

\section{Conclusion and Future Work}\label{Conclusion}
In this paper, we introduced a novel training algorithm using the bidirectional LSTM model to train the genome dataset. The obtained results of the training algorithm are fixed-length vector representations of the reference genome which is called the trained model. We also explained how this trained model can be leveraged to learn locality-sensitive hashing (LSH) to perform LSH-based sequence alignment. We reported the results of a case study in which the proposed modeling is performed. As future works, we can conduct experimental results to evaluate the proposed sequence alignment method. In addition, the obtained trained model can apply on genome assembly problems.

\bibliographystyle{IEEEtran}
\bibliography{sample-base}{}
\end{document}